\documentclass{myThesisTemp}

\usepackage[outdir=./]{epstopdf}
\newcommand{\qsq}{$\mathrm{VQ}^{2}\!\mathrm{A}$\xspace}

\title{
Foundational Question Generation for Video Question Answering via an Embedding-Integrated Approach
}{

}

\author[korean]{오 주 영}{}{}
\author[english]{Oh}{Ju-Young}{}

\advisor{이 성 환}{Seong-Whan Lee}

\department{
Artificial Intelligence
}{
인 공 지 능 학 과
}

\committee[1]{Seong-Whan Lee} 
\committee[2]{Won-Zoo Chung}
\committee[3]{Tae-Eui Kam}

\graduateDate{2026}{2}
\submitDate{2025}{11}
\approvalDate{2025}{12}

\captionLineSpacing{150}
\abstractLineSpacing{200}
\krAbstractLineSpacing{200}
\TOCLineSpacing{200}
\contentLineSpacing{200}
\acknowledgementLineSpacing{200}

\begin{document}

\chapter{Introduction}

Video Question Answering (VQA) is a multimodal learning task~\cite{kim2024te} that integrates computer vision and natural language processing to enable the understanding and reasoning over dynamic video content. In this task, a model is required to comprehend the temporal evolution of scenes, identify relevant entities, and infer their relationships in order to correctly answer a given question. Unlike static image-based QA, VQA captures motion patterns~\cite{yang2007reconstruction, lee2025full}, object interactions, and temporal dependencies that unfold across multiple frames. This makes VQA one of the most comprehensive benchmarks for evaluating a model’s ability to perform cross-modal reasoning.

VQA has drawn increasing research attention due to its significance and wide range of real-world applications, including education, healthcare, and surveillance systems~\cite{lee2015motion, maeng2012nighttime}. Despite the rapid progress in this field and the development of large-scale video-language datasets, achieving an effective alignment between linguistic expressions and visual representations still remains a fundamental challenge. Nevertheless, recent studies have demonstrated substantial progress in this direction, with works such as~\cite{videocontextalignertransformer, blip2, lee2024text} achieving impressive results by learning context-aware alignment mechanisms that connect the two modalities more effectively.

\begin{figure}[!t]
\centering
\centerline{\includegraphics[width=\linewidth]{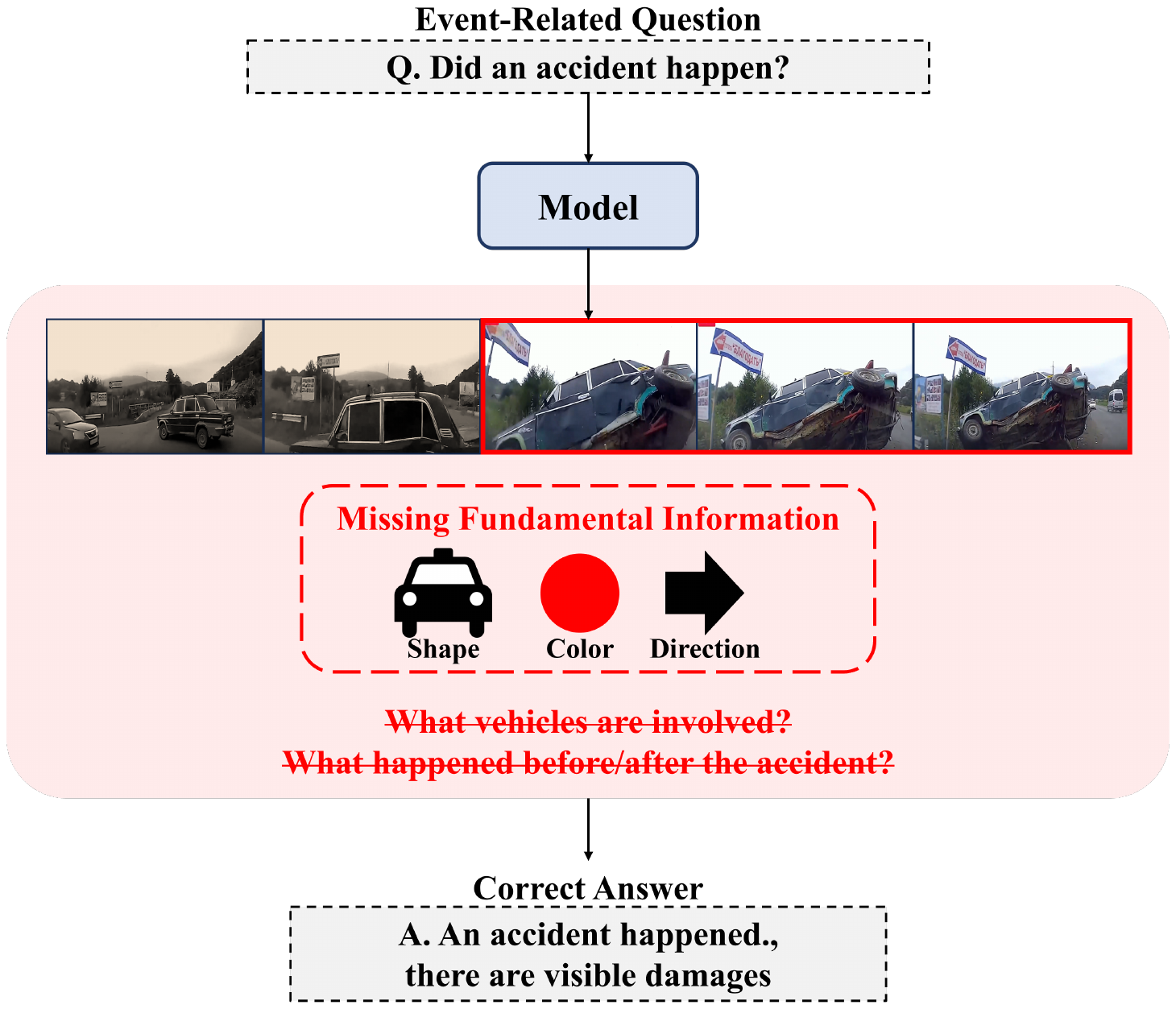}}
\caption{The existing dataset only focuses on event-centric information of video, but not on fundamental information of video such as shape, color, and direction of objects.}
\label{fig:issue}
\vspace{-0.4cm}
\end{figure}

Most existing VQA methods rely on CLIP-based encoders, which utilize the pretrained image–text alignment capability learned from massive image–caption pairs. Although video-specific encoders~\cite{videollava, VATT, slowfast, videomae} designed for spatio-temporal data are available, achieving strong visual–textual correspondence requires both pretrained visual and text encoders, which is an advantage that CLIP inherently provides. To further enhance multimodal reasoning, FrozenBiLM~\cite{frozenBiLM} introduces a lightweight connector that bridges the frozen CLIP image encoder and a bidirectional language model through masked language modeling, improving efficiency and interpretability. Similarly, ViLA~\cite{vila} presents QFormer-Distiller, a component that transfers representational knowledge~\cite{roh2007accurate} from BLIP’s Q-Former~\cite{blip} to strengthen cross-modal alignment.

Although CLIP provides powerful cross-modal representations~\cite{lim2000text}, its training on static images makes it heavily dependent on textual annotations to infer spatio-temporal relations. However, current VQA datasets are predominantly event-centric, describing actions~\cite{ahmad2006human} or outcomes while often neglecting basic visual properties such as object type, color, shape, and orientation. While event-centric annotations contain semantic cues, they represent only partial information, limiting the model’s ability to acquire a holistic understanding of a scene or its temporal evolution.

As depicted in Fig.~\ref{fig:issue}, a model trained exclusively on event-centric data often focuses narrowly on specific temporal moments, such as the instant of collision, while ignoring contextual information from earlier or later frames. This limitation arises because event-centric supervision alone does not provide enough descriptive details to fully answer complex questions. Consequently, such a model tends to exhibit weak generalization and limited high-level reasoning, as it lacks the broader contextual understanding required for causal inference~\cite{lee1997new, lee1999integrated, lee2020uncertainty}.

Moreover, even when the model successfully detects that an accident has occurred, it frequently fails to recognize which vehicles are involved, how the event unfolds over time, or what happens afterward. This incomplete understanding restricts the model’s ability to establish causal and temporal relations between frames. To overcome this issue, it is essential to incorporate fundamental scene attributes, including shape, color, and motion direction, which enable the system to consistently track objects and comprehend visual transitions over time. By integrating these attributes, the model can achieve a more comprehensive and coherent interpretation of the overall video context, extending beyond localized event detection.

To address these challenges, we propose FIQ (Fundamental Question Generation with Question Embedding Integration for Video Question Answering). The core idea of FIQ is to augment existing VQA datasets by automatically generating general question–answer pairs that emphasize the foundational understanding of visual elements such as object shape, color, and direction. This additional supervision encourages models to learn more complete video representations. Furthermore, we introduce the VQ-CAlign module, which integrates question embeddings as task-specific guidance to preserve the model’s awareness of the target question context. This design allows the model to maintain balanced attention between broad visual understanding and task-directed reasoning.

Our main contributions are summarized as follows:
\begin{itemize}
\item We propose FIQ, a framework that generates question–answer pairs to enrich the understanding of fundamental visual attributes, thereby improving the generalization and reasoning capability the model.
\item We design the VQ-CAlign module, which incorporates question embeddings to integrate task-specific information into the multimodal alignment process.
\item Our approach achieves state-of-the-art performance on the SUTD-TrafficQA dataset, outperforming previous baselines and validating the effectiveness of our framework.
\end{itemize}
\chapter{Related Works}
\section{Video Question Answering}
VQA is the task of interpreting the semantic information within a video to generate an appropriate answer to a given query. To reduce computational overhead, two main approaches have been investigated, namely adapter-based methods and text-based representation learning. Adapter-based methods~\cite{clip-adapter, vita, graphadapter} aim to minimize computational cost by enabling large language models (LLMs) to adapt to downstream tasks without the need for full fine-tuning. For example, Tem-Adapter~\cite{temAdapter} introduces an alignment method that leverages auto-regression to integrate semantic and temporal information~\cite{lee2001automatic} from the video domain into the image domain. While several studies employ textual information to enhance spatial and temporal understanding in videos, achieving competitive VQA performance using only textual representations remains uncommon. Vamos~\cite{wang2024vamos} presents a text-based video understanding framework that achieves strong performance without visual features by generating task-agnostic textual representations, demonstrating that leveraging textual data alone can improve results. Similarly, ColPro~\cite{cai2024empowering} integrates three distinct task-specific prompts to mitigate catastrophic forgetting during training. Despite differing objectives, both methods achieve remarkable results without relying on visual embeddings~\cite{min2022attentional}. Our approach aligns with these studies by utilizing textual prompts to enhance the interpretability of video data and capture its spatial and temporal characteristics. Furthermore, we employ both a language model (LM) and a large language model (LLM) to generate task-agnostic question–answer pairs that provide a comprehensive overview of the fundamental components of the video.

\section{General Question Generation}
Question Generation (QG) is the task of automatically producing questions with the goal of expanding semantic diversity and uncovering insights beyond explicitly visible information. QG methods are generally divided into two main categories. The first involves generating task-specific questions, and the second focuses on generating general questions. Task-specific QG has achieved state-of-the-art performance across multiple methods, as questions serve as one of the most informative inputs, guiding models on how to interpret data. Prophet~\cite{pllm} and SGSH~\cite{sgsh} propose knowledge-based question generation (KBQG) frameworks that produce natural language questions using external knowledge sources beyond the provided images. While QG effectively supports task-specific understanding, it serves as a powerful data augmentation technique by generating general questions regarding attributes such as color, object type, and quantity. \qsq~\cite{changpinyo2022all} demonstrates this by utilizing the T5 model to integrate multiple models and generate diverse, multilingual Q\&A pairs. Similarly, the All-in-One QAG model~\cite{ushio2023practical} emphasizes the potential of textual captions~\cite{lee1990translation} to enrich VQA datasets by incorporating details not explicitly represented in visual content. Building upon these approaches, we employ both LMs and LLMs to generate contextually rich and temporally informed questions, providing a more adaptable and robust framework for question generation.

\chapter{Methods}

\begin{figure*}[t!]
\begin{center}

\includegraphics[width=\linewidth]{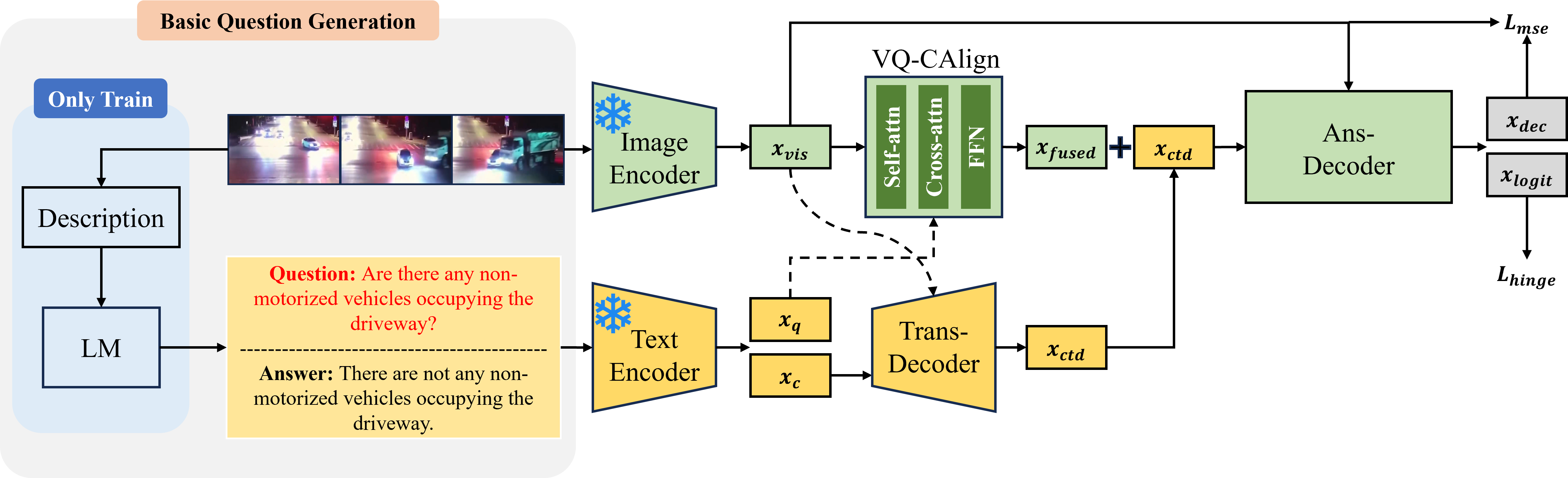}

\caption{Overall architecture of FIQ. It consists of four pivotal sub-processes. Q\&A pair which contains the general information of video first generated using language model such as T5~\cite{t5}, and GPT~\cite{gpt4}. The frozen text encoder takes these generate Q\&A pairs with the original dataset as an input, and each question embeddings and answer candidate embeddings are passed to the Trans-Decoder and VQ-CAlign. The frozen image encoder takes video data as input, and extracted visual features are passed to VQ-CAlign with question embeddings. Both modalities are merged, and passed to the Ans-Decoder, which fuses visual and textual information to align the temporal information.}

\label{fig:main}
\end{center}
\end{figure*}

FIQ consists of four main processes: Fundamental question generation, textual representation refinement, integration of question embeddings, and visual representation alignment. Fig.~\ref{fig:main} shows the overall framework of FIQ. The following sections present a detailed description of each component in the process.

\section{Preliminaries} 
The goal of the multi-choice VQA task is to determine the most appropriate answer $a_{final}$ from the provided options, given the question $x_{q}$ and the visual feature $x_{vis}$. For each answer candidate $x_{c}$, a corresponding score is computed, and the candidate with the highest score is selected as the final answer. The predicted answer $\hat{a}_{final}$ is obtained as follows:

\begin{equation} \label{eq:vqa}
\hat{a}_{final} = \text{argmax}(x_{c}|x_{vis},x_{q}).
\end{equation} 

\section{Fundamental Question Generation}\label{qnaGenmethod}

We employ VideoChat2~\cite{mvbench} to generate comprehensive textual descriptions that capture both low-level visual features, such as color, texture, and object appearance, and high-level semantic information, including motion patterns and temporal event order. These descriptions provide a balanced representation of static and dynamic aspects of the video content. After obtaining the initial textual outputs, we apply a filtering process to remove repetitive or irrelevant numerical expressions that do not accurately reflect the given video context. This refinement ensures that only meaningful and contextually appropriate descriptions are preserved for subsequent question generation. From the filtered descriptions, we utilize LMs, specifically T5 and GPT-4o-mini, to construct corresponding Q\&A pairs. Following the approach of \qsq~\cite{changpinyo2022all}, our framework is designed to guide the model through three key stages: candidate answer extraction, question generation, and answer validation. This structured pipeline promotes the generation of high-quality and contextually relevant question–answer pairs that capture both fine-grained and global video semantics.

\subsection{Candidate Answer Extraction}

In the candidate answer extraction stage, the language model is guided to identify potential answer candidates that represent the essential semantic elements of the video descriptions. These candidates include a variety of linguistic forms, such as noun phrases, named entities, and short open-class word sequences that describe objects, scenes, or actions. Additionally, the extraction process includes boolean literals (e.g., \textit{yes} or \textit{no}) and quantitative expressions that indicate object counts, explicitly including zero when no relevant entity is present. This comprehensive approach ensures that the candidate answers encompass both categorical and numerical information, thereby enabling a more balanced and diverse foundation for subsequent question generation. By systematically covering these answer types, the model reflects a broader range of semantic information from the video, which contributes to a richer representation and a more diverse set of question contexts.

\subsection{Question Generation}

During the question generation phase, the language model rewrites each source sentence containing a candidate answer into a well-formed interrogative expression. To encourage a broad range of question patterns, we instruct the model to generate diverse question types, including but not limited to formulations beginning with ``How many'', ``Where is'', and ``Is there''. This diversity ensures that the generated questions assess different aspects of visual and temporal understanding, ranging from quantitative reasoning to spatial localization and existence verification. All generated question–answer pairs are constrained to remain within 77 tokens to comply with the maximum input length of the pretrained CLIP text encoder. This restriction maintains compatibility with downstream multimodal processing while preserving linguistic clarity and informativeness.

\subsection{Answer Validation}
To ensure the reliability and consistency of the generated Q\&A pairs, we apply a token-level F1 score~\cite{tokenlevel} as an evaluation metric. It verifies whether each candidate's answer accurately corresponds to the meaning of the original descriptive sentence. This metric provides a fine-grained validation of lexical and semantic similarity while minimizing the inclusion of semantically inconsistent samples. When the computed F1 score falls below a threshold value of 0.54, the corresponding sample is excluded from the dataset to maintain high-quality and contextually faithful pairs. Each validated question is paired with a single correct answer, forming a one-to-one correspondence between the question and its associated answer. However, since the SUTD-TrafficQA dataset adopts a multi-choice format, each question must include multiple answer options to ensure compatibility with the dataset’s structure. To achieve this, we construct the positive answer using the ground-truth information derived from the target video ID, while the remaining negative answers are sampled from unrelated video IDs. This approach ensures that the negative answers are semantically diverse yet contextually plausible.  

To introduce sufficient randomness and reduce potential bias, we randomly select three distinct video IDs from the dataset and choose one answer from each of their available answer sets to serve as the negative options. This process encourages variability and prevents the model from overfitting to repetitive or overly similar distractors. Finally, the validated and reformatted Q\&A pairs are integrated into the original SUTD-TrafficQA dataset, thereby expanding its diversity and improving the overall robustness of the training data for the downstream VQA model.

\subsection{Textual Representation Refinement} \label{textproc}

To enhance the quality and discriminative capability of textual embeddings, we adopt a textual refinement process that focuses on extracting semantically rich and task-relevant representations from the input text. Specifically, we utilize a frozen CLIP~\cite{clip} text encoder in conjunction with a Transformer-based decoder (Trans-Decoder) to process both the question and the corresponding candidate answers. 

Given a question–answer pair, the textual encoder produces a sequence of token embeddings that capture syntactic structure and semantic meaning. Each question and its four candidate answers are encoded into distinct embedding vectors, ensuring consistent alignment across all options. In parallel, we extract visual representations from a frozen CLIP image encoder, denoted as $x_{vis} \in \mathbb{R}^{N \times D}$, where $N$ represents the number of sampled video frames and $D$ denotes the feature dimension of the encoder. These visual embeddings provide complementary spatial and temporal context for the textual representations.  

Furthermore, the textual embeddings of answer candidates are denoted as $x_{c} \in \mathbb{R}^{T \times D}$, where $T$ corresponds to the token sequence length of each textual input. The Trans-Decoder is then applied to integrate these two modalities, receiving both $x_{vis}$ and $x_{c}$ as inputs to produce refined candidate embeddings, denoted as $x_{ctd} \in \mathbb{R}^{T \times D}$. This process effectively aligns the semantic space between visual and textual modalities by attending to relevant visual cues while refining linguistic features that contribute to the answer prediction.  

By leveraging frozen encoders, the framework retains the robustness of pretrained multimodal knowledge while ensuring computational efficiency. The Trans-Decoder selectively adapts the textual embeddings toward video-relevant semantics without updating the parameters of the base encoders. This design allows the system to capture intricate correspondences between the question, answer candidates, and the underlying visual scene.

\subsection{Integration of Question Embeddings}\label{qalign} 

To effectively capture the temporal and spatial dynamics inherent in video data, we introduce learnable positional embeddings that explicitly encode frame-wise positional information. These embeddings serve to enrich the visual representations by incorporating sequential dependencies that occur across frames, thereby enabling the model to maintain temporal coherence when interpreting motion or event progressions within a scene. Such positional embeddings help the model distinguish between subtle temporal variations, such as the transition between object states or the continuation of an action. The formulation is expressed as follows:

\begin{equation} \label{eq:learnable}
x_{vpe} = x_{vis} + e_{pos}, 
\end{equation}

\noindent where $e_{pos}\in \mathbb{R}^{N \times D}$ represents the learnable positional embedding, and $x_{vpe} \in \mathbb{R}^{N \times D}$ denotes the visual feature enriched with positional information.

Although general Q\&A pairs enhance the model’s ability to capture the fundamental semantics of video content, the inclusion of task-specific information remains essential for achieving a more accurate and goal-oriented understanding. To bridge this gap, we propose the VQ-CAlign module, which fuses question embeddings with visual representations to inject linguistic intent directly into the visual reasoning process. The module consists of three primary components: self-attention, cross-attention, and a feedforward network. It takes as input both visual embeddings $x_{vis}$ and question embeddings $x_{q}\in \mathbb{R}^{T \times D}$, and the fusion process is defined as:

\begin{equation} \label{eq:qatten}
x_{fused} = \text{VQ-CAlign}(x_{vpe}, x_{q}).
\end{equation}

\noindent Within the VQ-CAlign module, the self-attention mechanism first operates solely on the visual embeddings $x_{vpe}$. It uses these embeddings as the query, key, and value to compute internal correlations among video frames. This step enables the model to identify long-range temporal relationships and local spatial dependencies within the visual sequence. Through this operation, the model learns which frames or visual regions are contextually related, such as the correlation between an object’s earlier and later states. The output of this stage, denoted as $x_{self}\in \mathbb{R}^{N \times D}$, contains temporally consistent and contextually enriched visual information that captures the internal coherence of the video.

The next component, the cross-attention module, plays a critical role in linking visual and linguistic information. Here, $x_{self}$ serves as the query, while the question embedding $x_{q}$ acts as the key and value. This interaction allows the model to selectively attend to visual regions that are semantically relevant to the given question. For example, if the question refers to “the color of the vehicle,” the module focuses its attention on frames and spatial regions containing the mentioned object, while down-weighting unrelated background elements. As a result, the cross-attention mechanism effectively injects linguistic guidance into the visual features, producing $x_{ca}\in \mathbb{R}^{N \times D}$ that integrates both visual evidence and question-specific semantics. This process enables the model to reason not only about visual appearance but also about the contextual meaning implied by the question.

Following the attention stages, the feedforward network further refines the cross-attended features by applying nonlinear transformations to enhance representational richness and filter out redundant information. The resulting feature, $x_{fused}\in \mathbb{R}^{N \times D}$, represents a harmonized fusion of the visual and question embeddings, encapsulating both the structural and semantic aspects of the video in a task-aware manner.

To reinforce task-specific alignment, we combine the fused feature $x_{fused}$ with the textual embeddings $x_{ctd}$ obtained from the Trans-Decoder. This step ensures that the model retains the detailed textual cues learned from the question–answer structure while integrating them with visual understanding. The combination is formulated as:

\begin{equation} \label{eq:mix}
x_{mix} = x_{fused} + x_{ctd},
\end{equation}

\noindent where $x_{mix}\in \mathbb{R}^{N \times D}$ denotes the final integrated representation that encodes both the visual–linguistic correspondence and task-specific information. This fusion allows the model to interpret visual context with respect to the question’s intent, ultimately improving reasoning accuracy and consistency in the VQA task.

\chapter{Experiments}

\section{Setup}

\subsection{Hyperparameters}

During preprocessing, we employ CLIP~\cite{clip} with a ViT/B-16 backbone, setting the visual feature dimension to 512. Each video is divided into eight clips, and from each clip, we extract 16 consecutive frames, resulting in a total of 128 frames per video. For training, we configure the batch size to 32 and the number of epochs to 37. We adopt an exponential moving average (EMA) with a decay rate of 0.9999. We employ a cosine decay learning rate schedule with a decay factor of 2. Additionally, we incorporate a learnable embedding layer with a dropout rate of 0.2 and a maximum sequence length of 128. For all attention-based modules, the number of attention heads is set to 16.

\subsection{Dataset}
We conduct experiments on the SUTD-TrafficQA dataset, a large-scale benchmark specifically designed for evaluating reasoning and understanding in traffic-related scenarios. The dataset contains 10,080 video clips and 62,535 human-annotated question–answer pairs, providing a rich set of multimodal cues for assessing both perception and reasoning capabilities. Unlike generic VQA datasets, SUTD-TrafficQA emphasizes real-world traffic environments, requiring models to reason about object interactions, accident causes, and temporal dependencies across scenes.

SUTD-TrafficQA includes six distinct reasoning tasks, each representing a different cognitive aspect of video understanding in traffic domains:

\noindent\textbf{Basic Understanding (B).}
This task evaluates the model’s ability to perform fundamental interpretation of traffic scenes, such as identifying objects, recognizing actions, and performing event classification or counting. It focuses on direct perception-based reasoning that requires minimal temporal inference.

\noindent\textbf{Event Forecasting (F).}
This task assesses the model’s capability to predict potential future events based on the current situation. Given a partial video and a question, the model infer possible outcomes, such as predicting collisions or traffic violations that are likely to occur.

\noindent\textbf{Reverse Reasoning (R).}
In this task, the model is required to infer the preceding events that might have caused the current situation in a given video segment. It evaluates the temporal reasoning ability of the model to reconstruct event sequences in reverse order.

\noindent\textbf{Introspection (I).}
This task measures the model’s capacity for preventive reasoning, in which it provides advice or identifies actions that could have prevented an accident.

\noindent\textbf{Attribution (A).}
This task focuses on identifying the underlying causes of traffic events. The model determines the most plausible factor responsible for an outcome, such as whether an accident occurred due to speeding, obstruction, or violation of traffic signals, among several answer candidates.

\noindent\textbf{Counterfactual Inference (C).}
This task differs from the others as it requires reasoning over hypothetical or imaginary situations not explicitly shown in the video. The model reasons about what would have happened under alternate conditions, which reflects high-level reasoning beyond direct visual evidence.

Together, these six tasks comprehensively evaluate a model’s understanding of both concrete visual information and abstract causal reasoning in complex, real-world traffic environments. They collectively test perception, temporal comprehension, counterfactual reasoning, and predictive inference—key elements necessary for robust video question answering.

\begin{table*}[!t]
\caption{
Performance comparison with state-of-the-art methods on SUTD-TrafficQA and each (H) and (H$^*$) represent training prompts with and without adapter heads. (H) and (A) represent methods for adding prompts, respectively. Avg represents an average accuracy for all six tasks.}

\centering
\begin{tabular}{l| c c c c c c c}
\toprule
\multirow{2.5}{*}{\textbf{Methods}} &\multicolumn{7}{c}{\textbf{SUTD-TrafficQA}} \\ 
\cmidrule(lr){2-8}
~ &\multicolumn{1}{c}{\textbf{B}} & \multicolumn{1}{c}{\textbf{F}} &  \multicolumn{1}{c}{\textbf{R}} &  \multicolumn{1}{c}{\textbf{C}} & 
\multicolumn{1}{c}{\textbf{I}} &  \multicolumn{1}{c}{\textbf{A}} &
\multicolumn{1}{c}{\textbf{Avg}}
\\
\midrule
Unsupervised CLIP~\cite{clip} & 25.6 & 20.1 &34.0   & 30.8 & 22.8 & 28.8 &  26.5   \\ 
CLIP~\cite{clip} + Template & 31.8 & 36.0 & 29.9 & 71.8   & 22.1 &  33.4 & 32.3   \\ 
Totally finetuning & 39.8 & 35.1 & 46.6 & 45.6 & 37.2  & 40.5  & 40.3   \\ 
Partially finetuning  & 41.6 & 37.8 &  44.6  & 50.0  & 33.1 & 41.7 & 41.7 \\ 
LoRA~\cite{lora} & 38.7 & 38.7 &  36.7  & 37.9  & 34.5 & 38.1 & 38.3  \\
CLIP-Adapter~\cite{clip-adapter} &  35.8 &  32.0 &  35.4 &  42.3 & 33.1   &32.1  &34.8    \\ 
Multi-layer Adapter~\cite{clip-adapter} &  30.5 &  26.6& 26.5 & 38.5 & 28.3  & 25.8 & 29.1   \\ 
Prompt learning (H)~\cite{coop} & 42.4  &  32.4  &   45.2  & 55.5  &  40.7 & 43.6   & 42.9   \\ 
Prompt learning (H$^*$)~\cite{coop} &  40.3 & 33.2   &  41.0   &  46.5  & 34.9  &   38.4 &39.7  \\
Prompt learning (A)~\cite{vpt} &  41.7  &  31.5 &  40.1 & 48.4  & 33.1  &  41.4 & 41.1 \\ 
Tem-Adapter~\cite{temAdapter} & 45.5 & 37.2 &45.8   & 54.5 & 35.1 & 48.3 &  46.0 \\
\midrule
FIQ &  \textbf{46.9} &   \textbf{43.5} &  \textbf{52.5} &    \textbf{54.0} &   \textbf{39.8}  &   \textbf{51.8} &  \textbf{48.4}    \\ 
\bottomrule
\end{tabular}
\label{table:base}
\end{table*}

\section{Main Results} 

Our goal is to generate Q\&A pairs that incorporate fundamental visual and semantic information from videos, ultimately enhancing the model’s capacity for deep reasoning and inference. Although the SUTD-TrafficQA dataset already includes a substantial number of Q\&A pairs that capture basic scene-level information, these existing examples are often limited in diversity and insufficiently cover the low-level visual properties that support higher-order reasoning. 

To address this limitation, we augment the dataset by generating additional Q\&A pairs using a language model (LM), focusing on fundamental attributes such as object type, spatial orientation, and temporal relations. This augmentation expands the representational diversity of the dataset and enriches the training signals for the model. The integration of these Q\&A pairs results in an overall enhancement of model performance, as summarized in Table~\ref{table:base}.

In comparison with other competitive methods, our proposed approach demonstrates consistent and significant performance gains across five of the six evaluation tasks in the SUTD-TrafficQA benchmark. The improvement is particularly shown in the Forecasting (F), Reverse Reasoning (R), Introspection (I), and Attribution (A) tasks. 

The observed improvements in these tasks indicate that the generated Q\&A pairs successfully supply the missing foundational knowledge that enables the model to interpret and reason about factual sequences of events, object interactions, and visual dependencies over time. As a result, the model becomes more capable of connecting dynamic scenes with the linguistic cues presented in the questions, leading to more coherent and contextually grounded answers.

The experimental findings further reveal that our generated Q\&A pairs provide essential complementary information to the existing dataset. Although SUTD-TrafficQA already contains a subset of questions related to object-level and event-level understanding, these were not sufficient to capture the complete spatio-temporal structure needed for complex reasoning. The inclusion of our LM-generated pairs significantly enhances the diversity and granularity of such information. Importantly, even though our generation process was designed to produce questions focusing on fundamental visual features, the resulting pairs naturally incorporate temporal and causal cues due to the contextual nature of the extracted video descriptions. As discussed in Section~\ref{qnaGenmethod}, these descriptions inherently embed motion, order, and duration of events, enabling the language model to produce questions that reflect not only static attributes but also evolving dynamics within the scene. Consequently, the integrated dataset strengthens the model’s capacity to understand both what happens in the video and how and why those events unfold.

These results collectively demonstrate that augmenting the dataset with LM-generated fundamental Q\&A pairs is highly beneficial for tasks requiring factual inference and spatio-temporal reasoning. The improvements observed in F, R, I, and A confirm that the model develops a more grounded understanding of the causal and sequential nature of events. The enhanced reasoning performance suggests that the additional Q\&A pairs help the model capture implicit relationships among frames and maintain consistency across temporal boundaries. In contrast, the Counterfactual Inference (C) task exhibits relatively minimal change in performance, which can be attributed to its distinctive objective. Unlike the other tasks, which depend heavily on observable evidence within the video, task C requires reasoning about hypothetical scenarios that extend beyond the visual context. As such, the fundamental Q\&A pairs, designed primarily to reinforce factual and event-based understanding, contribute less to this form of speculative reasoning. Nevertheless, the stable performance in task C implies that our augmentation does not interfere with the model’s ability to generalize to abstract or counterfactual reasoning domains.

In summary, the empirical evidence highlights that the proposed integration of fundamental Q\&A pairs serves as an effective enhancement strategy, improving the reasoning depth and interpretability of multimodal models. By providing diverse, contextually rich, and temporally informed examples, our approach bridges the gap between surface-level visual recognition and deeper causal comprehension, paving the way for more robust and explainable video question answering systems.

\section{Ablation Studies}

To demonstrate the effectiveness and contribution of each component in our proposed framework, we perform an ablation study that systematically evaluates the impact of key components. Table~\ref{table:ablation} summarizes the performance improvements achieved by incrementally adding each component to the baseline model. 

We first evaluate the contribution of the VQ-CAlign module, which is introduced to integrate question embeddings as task-specific guidance features. The module fuses question and visual representations through cross-attention, enabling the model to align visual regions with the semantic focus of the question more effectively. Compared with the baseline Tem-Adapter~\cite{temAdapter}, our model incorporating VQ-CAlign shows a meaningful improvement in accuracy across all reasoning tasks. This improvement demonstrates that enriching the multimodal fusion process with question-aware information allows the model to better capture fine-grained correlations between linguistic cues and dynamic video events. The results also indicate that this attention-based integration strengthens temporal coherence and enhances interpretability by guiding the model toward question-relevant visual features during the reasoning process.

\begin{table*}[!t]
\caption{Ablation studies on the SUTD-TrafficQA by adding the VQ-CAlign and the dataset generated by T5 and GPT. Avg represents an average accuracy for all six tasks.}

\centering
\begin{tabular}{l| c c c c c c c}
\toprule
\multirow{2}{*}{\textbf{Methods}} &\multicolumn{7}{c}{\textbf{SUTD-TrafficQA}}\\ 
\cmidrule(lr){2-8}
~ &\multicolumn{1}{c}{\textbf{B}} & \multicolumn{1}{c}{\textbf{F}} &  \multicolumn{1}{c}{\textbf{R}} &  \multicolumn{1}{c}{\textbf{C}} & 
\multicolumn{1}{c}{\textbf{I}} &  \multicolumn{1}{c}{\textbf{A}} 
&\multicolumn{1}{c}{\textbf{Avg}}
\\
\midrule
Tem-Adapter~\cite{temAdapter} & 45.5 & 37.2 &45.8   & 54.5 & 35.1 & 48.3 &  46.0  \\ 
VQ-CAlign & 44.8 & 46.1 & 47.1 & 51.3 & 33.7  & 50.1  & 46.3  \\ 
\midrule
VQ-CAlign + T5~\cite{t5}  &   \textbf{46.1} &  \textbf{47.0} &   \textbf{52.1} &   \textbf{58.3}  &   \textbf{35.8} &  \textbf{50.9} & \textbf{47.8}  \\ 
VQ-CAlign + GPT~\cite{gpt4} &  \textbf{46.9} &   \textbf{43.5} &  \textbf{52.5} &   \textbf{54.0} &   \textbf{39.8}  &   \textbf{51.8} &  \textbf{48.4} \\ 
\bottomrule
\end{tabular}
\label{table:ablation}
\end{table*}

Beyond architectural optimization, we further evaluate the effectiveness of data augmentation through the addition of generated Q\&A pairs containing fundamental visual information. These Q\&A pairs are designed to supplement the original dataset with instances that emphasize low-level attributes such as object category, color, and spatial relations, thereby reinforcing the model’s understanding of visual fundamentals. To assess the contribution of language models in this process, we generate two separate sets of Q\&A pairs using T5 and GPT-based models, respectively. The inclusion of these generated pairs leads to notable improvements in model performance, indicating that providing more detailed and diverse textual supervision helps the model generalize across varied reasoning contexts.

When comparing the two language model configurations, we observe that while the T5-generated Q\&A pairs lead to moderate gains, their impact is limited by the model’s relatively constrained linguistic representation capability and reliance on smaller-scale pretraining data. In contrast, the Q\&A pairs generated using GPT exhibit substantially higher accuracy, achieving an overall performance of 48.4\%, which represents the best result among all evaluated settings. This outcome highlights the advantage of LLMs in producing semantically rich and contextually coherent questions that effectively capture the primary attributes of video data. The superior performance of GPT-based pairs suggests that the broader contextual understanding and stronger reasoning ability inherent in LLMs enable the generation of questions that more closely align with the visual and temporal characteristics of the video scenes.

\begin{figure}[!t]
\centering
\includegraphics[width=0.8\linewidth]{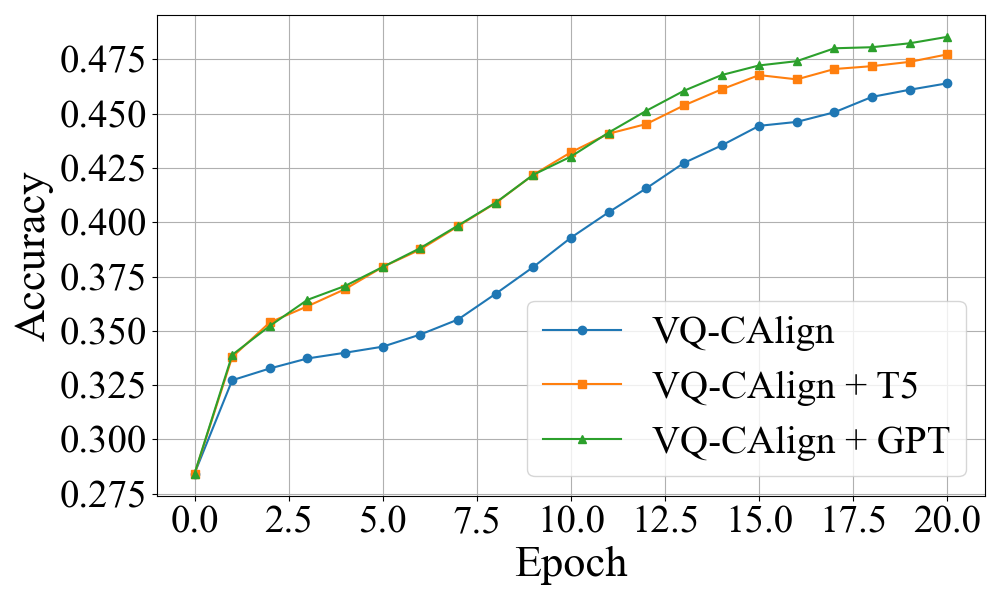}
\caption{Comparison between different LM-based Q\&A generation (T5, GPT) methods on SUTD-TrafficQA.}

\label{fig:accGraph}
\vspace{-0.4cm}
\end{figure}

Fig.~\ref{fig:accGraph} visualizes the overall accuracy improvements achieved through the sequential integration of each module and the generated Q\&A pairs. As depicted, all three FIQ configurations show consistent performance gains and rapid convergence, with all models stabilizing around epoch 20. Collectively, the ablation results confirm that both the VQ-CAlign module and the incorporation of LM-generated Q\&A pairs play complementary roles in improving model robustness, generalization, and reasoning efficiency in video question answering tasks.

\chapter{Conclusion}
In this paper, we propose FIQ, a framework that enhances video reasoning through a fundamental Q\&A pair generation method and VQ-CAlign mechanisms. Our approach produces foundational question–answer pairs to support event-centric textual annotations, leveraging LMs to strengthen the model’s reasoning capability and generalization performance. Furthermore, the VQ-CAlign module incorporates task-specific knowledge by question embedding representations, which allows the model to better handle downstream VQA tasks. Experimental results demonstrate that our method significantly improves the accuracy on reasoning-related tasks, confirming that integrating general video knowledge effectively boosts the model’s interpretive ability compared to existing approaches. In the future, we plan to develop a new dataset that embeds question information directly as potential answer candidates, aiming to further enhance reasoning consistency and contextual understanding.

\end{document}